\documentclass{article}

\usepackage[final]{neurips_distshift_2021}

\usepackage[utf8]{inputenc} 
\usepackage[T1]{fontenc}    
\usepackage{hyperref}       
\usepackage{url}            
\usepackage{booktabs}       
\usepackage{amsfonts}       
\usepackage{nicefrac}       
\usepackage{microtype}      
\usepackage{xcolor}         

\usepackage{graphicx}
\usepackage{subcaption}
\usepackage{multicol}
\usepackage{amsmath}
\usepackage{cleveref}
\usepackage{natbib}

\title{On The Reliability Of Machine Learning Applications In Manufacturing Environments}

\author{%
  Nicolas Jourdan \\
  TU Darmstadt, Germany\\
  \texttt{n.jourdan@ptw.tu-darmstadt.de} \\
  \And
  Sagar Sen \\
  SINTEF, Norway \\
  \texttt{sagar.sen@sintef.no} \\
  \And
  Erik Johannes Husom \\
  SINTEF, Norway \\
  \texttt{erik.husom@sintef.no} \\
  \And
  Enrique Garcia-Ceja \\
  SINTEF, Norway \\
  \texttt{e.g.mx@ieee.org} \\
  \And
  Tobias Biegel \\
  TU Darmstadt, Germany \\
  \texttt{t.biegel@ptw.tu-darmstadt.de} \\
  \And
  Joachim Metternich \\
  TU Darmstadt, Germany \\
  \texttt{j.metternich@ptw.tu-darmstadt.de} \\
}

\begin{document}

\maketitle
\begin{abstract}
      The increasing deployment of advanced digital technologies such as Internet of Things (IoT) devices and Cyber-Physical Systems (CPS) in industrial environments is enabling the productive use of machine learning (ML) algorithms in the manufacturing domain.
      As ML applications transcend from research to productive use in  real-world industrial environments, the question of reliability arises.
      Since the majority of ML models are trained and evaluated on static datasets, continuous online monitoring of their performance is required to build reliable systems.
      Furthermore, concept and sensor drift can lead to degrading accuracy of the algorithm over time, thus compromising safety, acceptance and economics if undetected and not properly addressed.
      In this work, we exemplarily highlight the severity of the issue on a publicly available industrial dataset which was recorded over the course of 36 months and explain possible sources of drift.
      We assess the robustness of ML algorithms commonly used in manufacturing and show, that the accuracy strongly declines with increasing drift for all tested algorithms. We further investigate how uncertainty estimation may be leveraged for online performance estimation as well as drift detection as a first step towards continually learning applications. The results indicate, that ensemble algorithms like random forests show the least decay of confidence calibration under drift.
\end{abstract}

\section{Introduction and Motivation}
Increasing digitization and the deployment of advanced technologies in the context of Internet of Things (IoT) and Industry 4.0 are transforming  manufacturing lines into Cyber-Physical Systems (CPS) that generate large amounts of data. The availability of this data enables a multitude of applications, including the development and deployment of ML algorithms for use cases such as condition monitoring, predictive maintenance and quality prediction \cite{jourdan_machine_2021}.
As ML applications are deployed to productive usage, their continuous reliability has to be guaranteed to protect human operators as well as the financial investments involved.
Manufacturing environments are fast changing, highly dynamic and inherently uncertain which poses the requirement for ML applications to be able to adapt to changing environments with reasonable effort and cost \cite{wuest2016machine}.
While the ability of adapting to a changing environment is often seen as a default property of machine learning \cite{wuest2016machine}, studies show, that the generalization ability of a model mainly depends on the configuration and variety of the available training data and is far from guaranteed \cite{chung2018unknown,jourdan2020identification}.
%
%
Long-term reliability and the handling of uncertainties caused by degrading equipment or faulty sensors are seen as key factors and major hurdles when deploying ML systems in manufacturing environments \cite{kusiak_smart_nodate,dependableAI2021}.
%
With respect to safety certification and risk assessment, online performance monitoring and uncertainty estimation are seen as critical for detecting drifts in the data distribution as well as estimating error magnitudes \cite{dependableAI2021}.
In the context of quality management, \cite{bretones_cassoli_frameworks_2021} showed, that the majority of the analyzed frameworks still lack any form of uncertainty estimation or online monitoring.

In this work, we analyze the long-term reliability of ML applications in the manufacturing industry, highlighting the domain-specific issues and potential sources of drift. We benchmark a set of ML algorithms that are most relevant in this domain for robustness to time-dependent drift on an industrial dataset. Further, we assess uncertainty estimation techniques and highlight their potential utility for online performance estimation and drift detection in the context of continual learning to overcome the issue of silently failing ML applications.
\newline
Similar experiments have been conducted in \cite{ovadia2019can} but did not consider non-deep learning algorithms, which are of high relevance in the manufacturing domain. Additionally, the introduced drifts were synthetic, while we evaluate on real-world time-dependend drifts. The performance degradation of a classifier (support vector machine) on the utilized dataset has been observed in \cite{vergara2012chemical}. In contrast to the existing work, we extend the analysis to a broad spectrum of commonly used ML algorithms and additionally analyze the expressiveness of uncertainty estimation techniques suitable for the respective algorithms.
\begin{figure}[t]
      \begin{subfigure}[t]{.45\linewidth}
            \centering
            \includegraphics[width=0.9\linewidth]{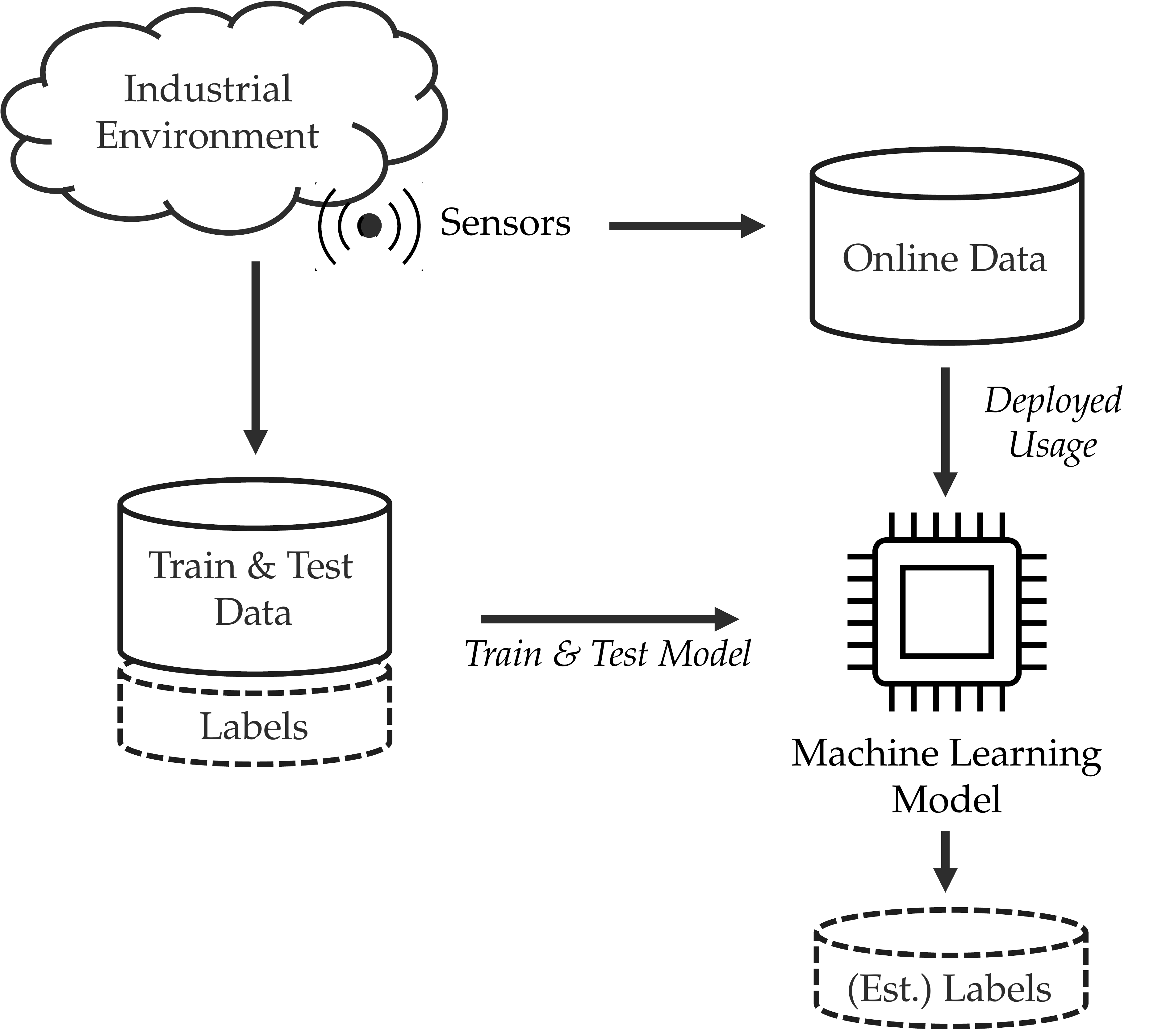}
            \caption{Stationary environment\label{subfig:stationary}}
      \end{subfigure}
      \begin{subfigure}[t]{.55\linewidth}
            \centering
            \includegraphics[width=0.9\linewidth]{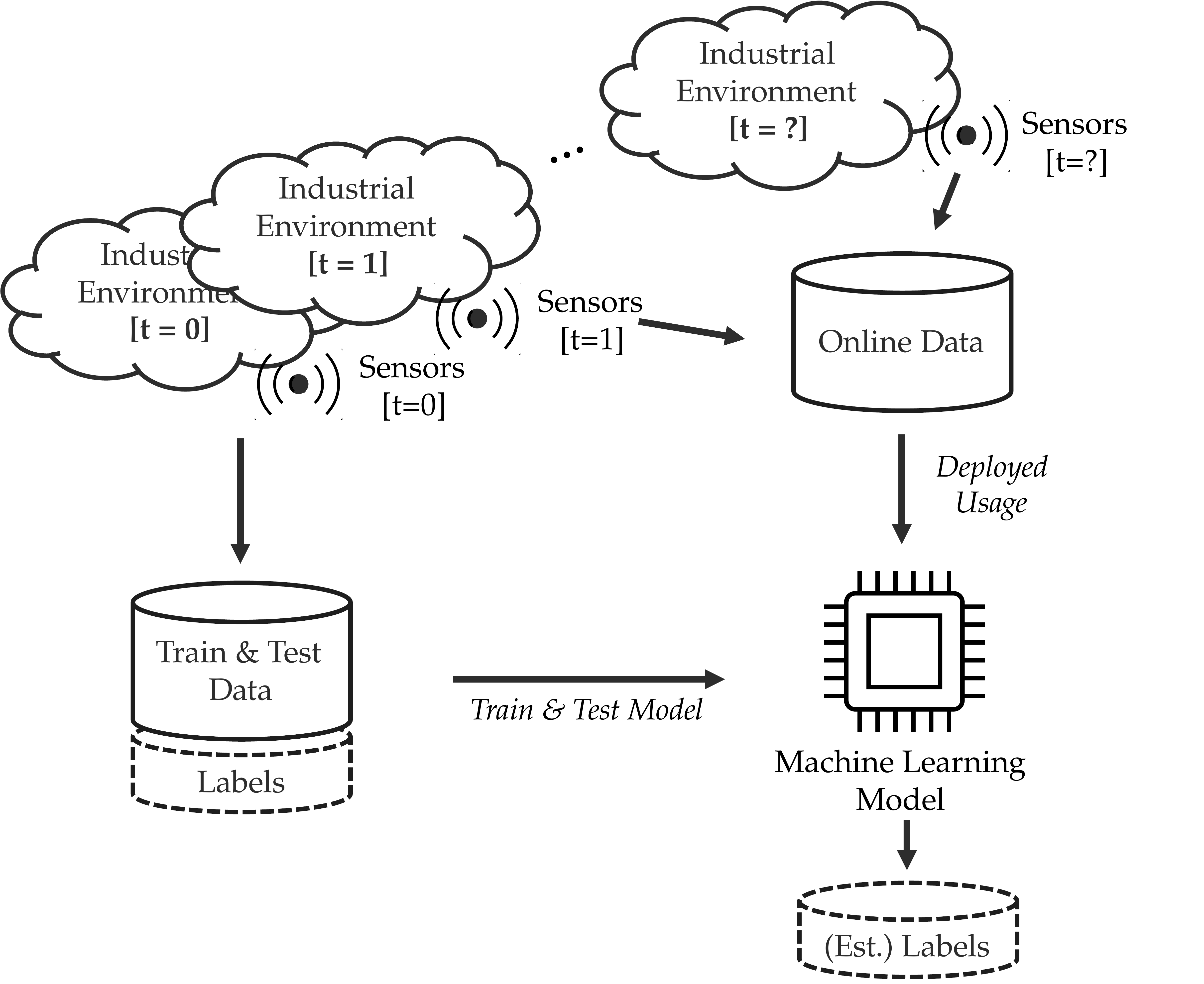}
            \caption{Environment under concept and sensor drift\label{subfig:drift}}
      \end{subfigure}
      \caption{
            Supervised machine learning in stationary conditions with static sensors (a) in contrast to a dynamic system where the environment as well as the sensors used for perception are non-stationary, which applies to the majority of manufacturing use cases of ML (b).
            In the latter case, static training/testing sets do not provide continuous performance guarantees.
            Figure based on \cite{vzliobaite2010adaptive}, adapted and extended with permission of the original authors.\label{fig:Drift}}
\end{figure}
\section{Methodology and Experiments\label{sec:problem}}
\textit{Concept drift} refers to a change in the underlying data distributions of machine learning applications.
Especially in the context of pattern recognition, the terms \textit{covariate shift} or \textit{dataset shift} are used interchangeably \cite{vzliobaite2016overview}.
Concept drift in the context of this publication, \textit{cf.} \Cref{fig:Drift} (b), can be defined as $P_{train}(\boldsymbol{X},Y) \neq P_{online,t}(\boldsymbol{X},Y)$,
where $P_{train}$ and $P_{online,t}$ denote the joint distributions of input samples $\boldsymbol{X}$ and target labels $Y$ during training and deployed usage of the model at time $t$, respectively.
\newline
Relevant short- and long-term sources of changes in manufacturing environments which may influence the reliability of ML models
include: Tool and machine wear \cite{kusiak_smart_nodate,dependableAI2021,tripathi2021ensuring,gao2017comprehensive}, changes in product configurations and material properties \cite{tripathi2021ensuring,gao2017comprehensive}, changes in upstream processes \cite{gao2017comprehensive}, changes in factory layout and machine placement \cite{sanz2012fuzzy}, differences in operator preferences and training \cite{tripathi2021ensuring, gao2017comprehensive}, seasons and time of day \cite{dependableAI2021}, environmental conditions such as temperature or humidity \cite{dependableAI2021,tripathi2021ensuring,gao2017comprehensive}, sensor failure/drift/recalibration \cite{kusiak_smart_nodate,dependableAI2021} and data transmission problems \cite{kusiak_smart_nodate}.
\newline
Reliable machine learning applications in dynamic environments may be established in two ways: Either the model and data acquisition setup employed are robust against the relevant sources of drift, e.g. \cite{jourdan_direct_tcm_2021}, or the model is continually assessed and, if required, adapted to the current environment in a continual learning setup \cite{diethe2019continual}.

Commonly, ML models with trained parameters $\boldsymbol{\theta}$ in classification tasks produce probability estimates $p\left(\hat{y}_c \mid \boldsymbol{x}, \boldsymbol{\theta}\right)$ for all classes $c\in\{1, \ldots, C\}$, given a sample of data $\boldsymbol{x}$. The probability may be used to assess the models' confidence/uncertainty in its own prediction. The confidence is referred to as \textit{well-calibrated}, when empirically, it is equal to the probability of the corresponding sample being correctly classified \cite{guo2017calibration}. Thus, confidence estimates that are well-calibrated even in the presence of concept drift, may be used to reason about the reliability of a ML model and determine if it should be adapted, i.e., retrained with new data. Additionally, well-calibrated confidence estimates may be used to identify product configurations or situations that are difficult for the ML model to handle. In the example of quality estimation, this could, e.g., trigger an additional human quality control for a given part or the manual inspection of a machine.
%
%
%
%
\subsection{ML Algorithms and Uncertainty Estimation Methods}
To maximize the value for practical applications, we assess ML algorithms that have been identified as most commonly used in the manufacturing domain by a recent review study \cite{fahle2020systematic}. We explicitly include non-deep learning algorithms in the analysis, as these are highly relevant in the manufacturing domain, where labeled datasets are often small. In the scope of this work, we focus on classification tasks. Implementation is done using \cite{scikit-learn} and \cite{tensorflow2015-whitepaper}. The hyperparameters are selected by performing a grid-search over the parameter space of the models, optimizing for accuracy. For each algorithm, we employ a confidence/uncertainty estimation method as described below:

\textbf{Support Vector Machine (SVM)} confidence estimates are obtained using Platt scaling \cite{platt1999probabilistic} of the sample distances to the separating hyperplane. The parameters for confidence estimation are fitted via 5-fold cross validation.

\textbf{Decision Tree (DT)} confidence estimates are computed as the fraction of training samples of the same class in the leaf node \cite{scikit-learn}.

\textbf{K-Nearest Neighbors (KNN)} confidence estimates are calculated similar to DTs. The probability of a class is computed as the fraction of training samples of the same class in the set of nearest neighbors, weighted by their distance.

\textbf{Random Forest (RF)} confidence estimates are computed as the mean predicted class probabilities of the trees in the forest. The individual tree confidences are computed as described above (DT). This method of confidence estimation for RFs has been shown to be superior to more complicated extensions \cite{bostrom2007rfcalib}.

\textbf{Neural Network}
For neural networks, we assess multiple recently proposed uncertainty estimation methods: \textbf{Max. Softmax Probability (NN)} \cite{hendrycks2016baseline}; \textbf{Deep Ensembles (NN-Ens)} \cite{lakshminarayanan2016simple} with $M=10$ ensemble members. Randomness is introduced by reshuffling of the training set as well as different random initialization for each NN in the ensemble; \textbf{Monte-Carlo Dropout (NN-MCD)} \cite{gal2016dropout} with $M=20$ forward passes for each sample. Dropout rate $p$ is set to $0.2$.

\subsection{Dataset}
For our experiments, we use the Gas Sensor Array Drift dataset \cite{vergara2012chemical} that was recorded at the University of California San Diego (UCSD). The dataset was recorded over 36 months at an industrial test rig. Due to the long recording time, the dataset contains both sensor drift due to aging sensors and concept drift due to external influences which resembles the expected environment conditions of a real-world ML application in manufacturing, \textit{cf.} \Cref{fig:Drift} (b). The dataset represents a classification task, in which the target variable is the type of gas (one of six) that is currently present in the apparatus. The experiments are perceived using 16 sensors and each row of the dataset contains 128 extracted statistical features (8 per sensor) of the corresponding experiment run with a total of 13,910 runs. The dataset is split into 10 consecutive batches, each capturing a varying amount of months.
\newpage
\subsection{Experiment Setting and Metrics}
We train all the models on a random $50\%$ split of the first $10$ months of the available data and use the remaining $50\%$ as the validation set for performance evaluation on non-drifted data. To be able to assess the robustness to drifts, we test on the remaining $26$ months. All available features are used for training and the experiments are repeated $10$ times with varying random seeds.

For evaluation, we employ two metrics capturing different aspects of interest:
\newline
\textbf{Accuracy $\uparrow$} is used to assess the performance of the model on the non-drifted test set as well as the performance degradation under drift.
The accuracy measures the percentage of correct classifications.
\newline
\textbf{Expected Calibration Error (ECE) $\downarrow$} \cite{naeini2015obtaining} is used to evaluate the calibration of the confidences produced by the model. The ECE
is closely related to calibration curves and
corresponds to the average gap between model confidence and achieved accuracy.
While the ECE has several shortcomings \cite{ovadia2019can}, we choose it over other calibration metrics for its simplicity and interpretability to strengthen the practical relevance.
%
\begin{figure}[t]
      \begin{subfigure}[t]{.48\linewidth}
            \centering
            \includegraphics[width=\textwidth]{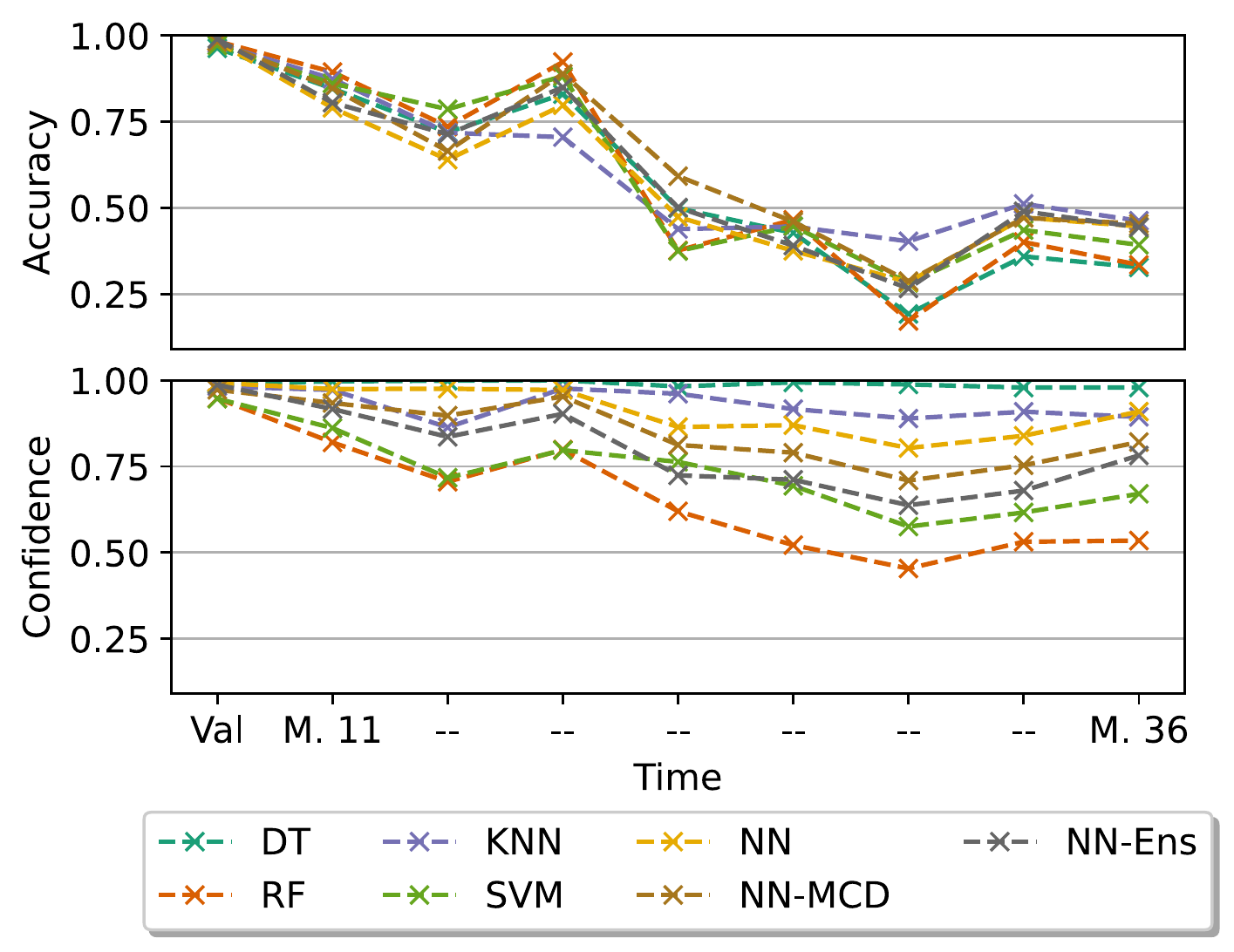}
            \caption{\textbf{Top:} Prediction accuracies over time. \textbf{Bottom:} Model confidences over time. \textit{Val} indicates the validation set containing non-drifted data.\label{subfig:result_a}}
      \end{subfigure}\hfill
      \begin{subfigure}[t]{.48\linewidth}
            \centering
            \includegraphics[width=\textwidth]{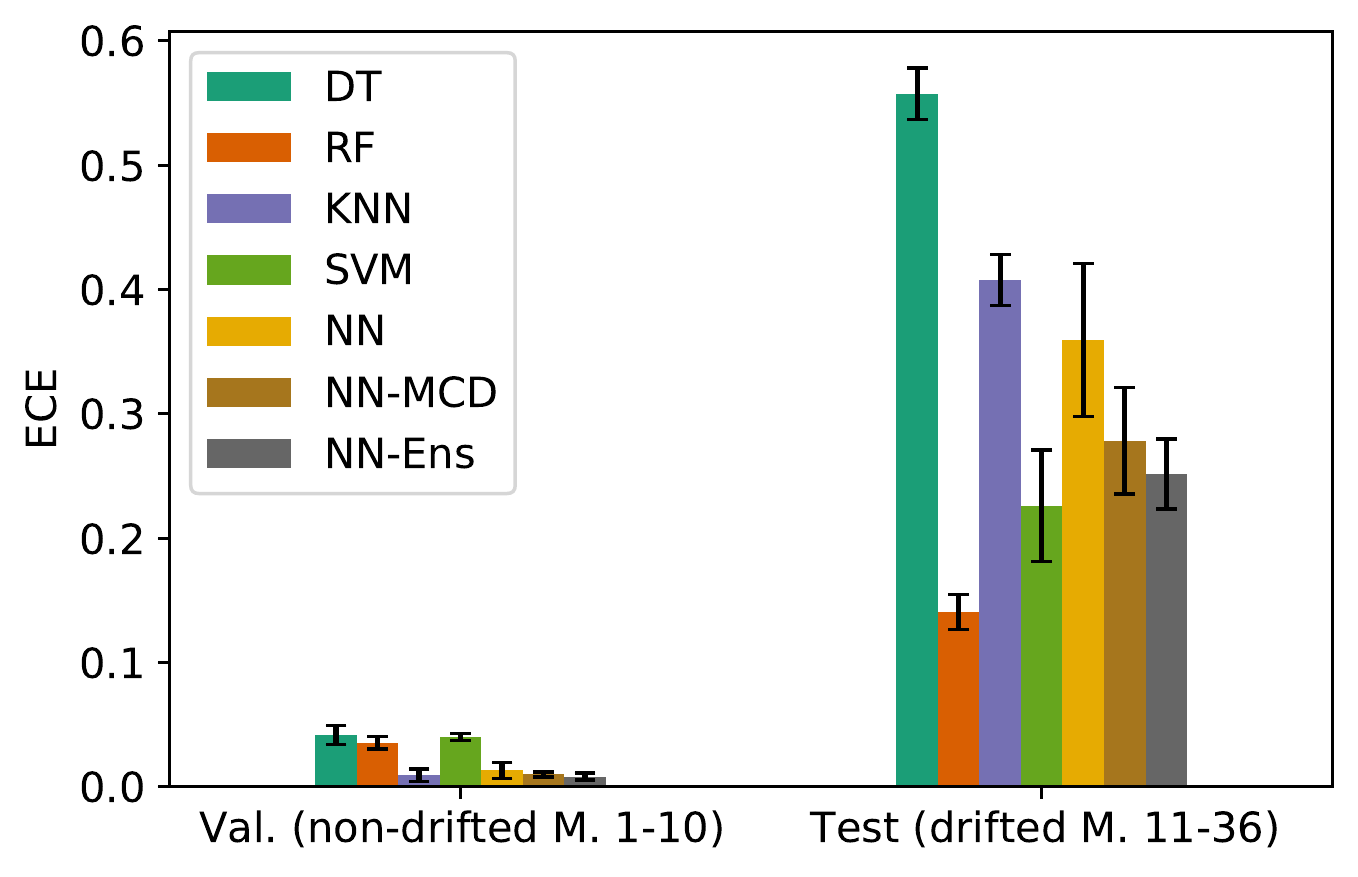}
            \caption{ECEs of the assessed models for the non-drifted validation set (months 1-10) as well as averaged over the drifted test set (months 11-36). Error bars show the standard deviation of the experiment runs.\label{subfig:result_b}}
      \end{subfigure}
      \setlength{\belowcaptionskip}{-10pt}
      \caption{Results of the experiments on the UCSD Gas Sensor Array \cite{vergara2012chemical} dataset regarding model accuracy, confidence and calibration under drift. All experiments are repeated $10$ times with different random seeds and the results consequently averaged.
            \label{fig:Results}}
\end{figure}
\subsection{Results\label{sec:experiments}}
As visualized in the upper part of \Cref{fig:Results} (a), the classification performance of all tested algorithms strongly degrades with an increasing time difference to the non-drifted validation set. This indicates, that none of the tested algorithms is robust against drifts in the environment. Thus, online monitoring and eventual model updates would be required to guarantee a reliable and safe application in real manufacturing environments.
In parallel, the reduction in accuracy is reflected by the lowering in confidence of a subset of the algorithms, most pronounced with RF.
These confidences may be used to identify drift in this scenario using frameworks such as \cite{baier2021detecting}. The calibration of the model confidences is further analyzed in \Cref{fig:Results} (b). Notably, all tested algorithms show well-calibrated confidences on the validation set, reflected by the low ECE, while the calibration strongly degrades for the drifted data. In addition, the error bars in \Cref{fig:Results} (b) show, that the standard deviation with respect to the ECEs of the $10$ experiment runs on the drifted data increased for all algorithms when compared to the standard deviation on the validation set. This further highlights, that the calibration of a model on drifted data, can usually not be inferred from the calibration on in-distribution data as it highly depends on the type and magnitude of the drift. Lastly, it can be observed that the calibration of the RF degrades least for the drifted data, followed by SVM and NN-Ens, supporting the visualization in \Cref{fig:Results} (a). Depending on the application requirements, the confidence of the RF may be used as a measure of the performance that can be expected of the algorithm as well as an indicator for drift. The observed comparably high calibration robustness of ensemble methods in the presence of drift aligns with previous work on deep ensembles \cite{ovadia2019can}. A possible reason may be, that each of the models in the ensemble slightly overfits on different aspects of the training data, which in combination yields lower confidences on the drifted data, as the predictions will deviate more strongly than on in-distribution data.
%
\section{Conclusion and Outlook\label{sec:conclusion}}
In this work, we highlighted the general relevance, implications and possible sources of drifts affecting the continuous reliability of ML applications in the manufacturing domain. Using an industrial dataset, we exemplarily show, that none of the most commonly used ML algorithms in manufacturing are robust against drifts in the data distribution inflicted by the environment or the sensors used for perception thereof. A consequent analysis regarding the confidence calibration of the algorithms showed, that in the majority of cases, the calibration strongly degrades with the drift, rendering the confidences unexpressive. Positively, the confidence calibration of ensemble algorithms such as random forests degrades less strongly and may be used to estimate the current performance and identify drifts. In a continual learning setup, the confidence could thus be used as a trigger signal for data collection and retraining of the respective model.

There are multiple opportunities for future work on drift detection and adaption through means of continual learning or domain adaptation specific to ML use cases in manufacturing such as condition monitoring, predictive maintenance or quality prediction to enable further adaption of ML applications in this domain. Especially the practical implementation of such systems on the shop floor level is still an open research issue.
\begin{ack}
      This project has received funding from the European Union’s Horizon 2020 Research and Innovation Programme under grant agreement No. 958357 (InterQ), and it is an initiative of the Factories-of-the-Future (FoF) Public Private Partnership.
\end{ack}

\bibliographystyle{unsrt}
\bibliography{literature.bib}

\appendix
\section{Appendix}
\subsection{Hyperparameters used in the experiments}
All input features are standardized using statistics calculated on the training set. Implementation is done using \cite{scikit-learn} and \cite{tensorflow2015-whitepaper}. Only non-default values are reported here:
\\
\textbf{Neural Network} The results are reported for a fully connected neural network with $3$ hidden layers consisting of $32,16$ and $8$ units respectively. The hidden layers use Rectified Linear Unit (ReLU) activations, while the class scores are calculated using the softmax function in the final layer. Cross-entropy loss is used in training with the ADAM optimizer for $50$ epochs at a learning rate of $0.001$. \textbf{Deep Ensembles (NN-Ens)} utilize the same architecture for the ensemble models. $M=10$ models are used in the ensemble. No dropout layers are used in the default case. For \textbf{Monte-Carlo Dropout (NN-MCD)}, dropout layers are added after each of the first two hidden layers with $p=0.2$. $M=20$ forward passes are performed for each sample.
\textbf{Decision Tree (DT)} The decision tree is fitted with a maximum depth of $6$.
\textbf{Random Forest (RF)} The random forest uses $100$ single estimators in the ensemble.

\end{document}